\pdfoutput=1

\documentclass[11pt]{article}

\usepackage[]{emnlp2021}

\usepackage{times}
\usepackage{latexsym}
\usepackage{microtype}
\usepackage{graphicx}
\usepackage{booktabs}  
\usepackage{multirow}
\usepackage{subfigure}
\usepackage{verbatim}
\usepackage{CJKutf8} 
\usepackage{amssymb}
\usepackage{amsmath}

\usepackage{algorithm}
\usepackage{algorithmic}
\usepackage{makecell} 
\usepackage[T1]{fontenc}

\usepackage[utf8]{inputenc}

\usepackage{microtype}
\usepackage{color}
\title{FedKD: Communication Efficient Federated Learning\\ via Knowledge Distillation}

\author{Chuhan Wu$^1$~~Fangzhao Wu$^2$~~Lingjuan Lyu$^3$~~Yongfeng Huang$^1$~~Xing Xie$^2$\\
    $^1$Department of Electronic Engineering \& BNRist, Tsinghua University  \\
     $^2$Microsoft Research Asia, 
          $^3$Sony AI\\
  {\tt\{wuchuhan15, wufangzhao, lingjuanlvsmile\}@gmail.com} \\
  {\tt yfhuang@tsinghua.edu.cn, xingx@microsoft.com}
  }
\date{}

\begin{document}
\maketitle

\begin{abstract}
Federated learning is widely used to learn intelligent models from decentralized data.
In federated learning, clients need to communicate their local model updates in each iteration of model learning.
However, model updates are large in size if the model contains numerous parameters, and there usually needs many rounds of communication until model converges.
Thus, the communication cost in federated learning can be quite heavy.
In this paper, we propose a communication efficient federated learning method based on knowledge distillation.
Instead of directly communicating the large models between clients and server, we propose an adaptive mutual distillation framework to reciprocally learn a student and a teacher model on each client, where only the student model is shared by different clients and updated collaboratively to reduce the communication cost.
Both the teacher and student on each client are learned on its local data and the knowledge distilled from each other, where their distillation intensities are controlled by their prediction quality.
To further reduce the communication cost, we propose a dynamic gradient approximation method based on singular value decomposition to approximate the exchanged gradients with dynamic precision.
Extensive experiments on benchmark datasets in different tasks show that our approach can effectively reduce the communication cost and achieve competitive results.

\end{abstract}

\section{Introduction}

Privacy protection of user data is a very important issue~\cite{shokri2015privacy}.
Federated learning is a well-known technique to learn intelligent models from decentralized user data
~\cite{mcmahan2017communication}.
It has been widely used in various applications such as intelligent keyboard~\cite{hard2018federated}, personalized recommendation~\cite{qi2020privacy} and topic modeling~\cite{jiang2019federated}.

In federated learning, the private data is locally stored on different clients~\cite{yang2019federated}.
Each client keeps a local model and computes the model updates from its local data. 
In each iteration, the model updates from a number of clients are uploaded to a server, which aggregates the local model updates into a global one to update its maintained global model.
Then, the server distributes the global update to each client to conduct a local model update.
This process is iteratively executed for many rounds until the model converges.
In this framework, the server and clients need to intensively communicate the model updates.
However, the communication cost is enormous if the model is in large size, which hinders the applications of many powerful but large-scale models like BERT~\cite{devlin2019bert} to federated learning.

In this paper, we propose a communication efficient federated learning method based on knowledge distillation (\textit{FedKD}).
Instead of directly communicating the large models between the clients and server, in \textit{FedKD} a small student model and a large teacher model are distilled from each other, where only the student model is shared by different clients and learned collaboratively, which can effectively reduce the communication cost.
More specifically, each client maintains a large local teacher model and a local copy of a small student model that is shared among different clients.
We propose an adaptive knowledge distillation method to enable the local teacher and student to learn from both the local data on its client and the knowledge distilled from each other, where their distillation intensities are controlled by the correctness of their predictions. 
The local teacher model on each client is locally updated, while the local updates of the student models from different clients are uploaded to a central server, which aggregates these local updates into a global one.
The server further distributes the global update to different clients to update their local student models.
This process is iteratively executed until the student model converges.
In addition, to further reduce the communication cost when exchanging the student model updates, we propose a dynamic gradient approximation method based on singular value decomposition (SVD) to compress the communicated gradients with dynamic precision.
Extensive experiments on benchmark datasets for different tasks validate that our approach can effectively reduce communication costs in federated learning and meanwhile achieve competitive performance.

The contributions of this paper are as follows:
\begin{itemize}
    \item We propose a communication efficient federated learning approach based on knowledge distillation, which can achieve competitive results with much less communication cost.
    \item We propose an adaptive mutual knowledge distillation method to encourage teacher and student  to learn from each other and be aware of their prediction correctness. 
    \item We propose a dynamic gradient approximation method based on SVD for gradient compression with dynamic precision to further reduce communication cost.
    \item We conduct extensive experiments on benchmark datasets for different tasks to verify the effectiveness and efficiency of our approach.
\end{itemize}

\section{Related Work}

\subsection{Federated Learning}

Federated learning (FL)~\cite{mcmahan2017communication} is a privacy-aware technique to learn intelligent models from decentralized data storage, where the raw user data never leaves where it is stored.
It has been widely used in many applications like intelligent keyboard~\cite{hard2018federated}, personalized recommendation~\cite{lin2020fedrec,qi2020privacy}, topic modeling~\cite{jiang2019federated} and medical natural language processing~\cite{ge2020fedner}.
In federated learning, there are usually a number of user devices that locally keep the privacy-sensitive user data, and a server that coordinates these user devices for collaborative model learning.
Each user device contains a local model copy and computes the model update based on the local data.
The model updates from a certain number of user devices are uploaded to the server, which aggregates the local updates into a global one for updating its maintained global model.
The updated global model is further distributed to user devices to update their local model copies.
This process will be repeated until the model is fully trained.
Since the model updates usually contain much less private information~\cite{mcmahan2017communication}, federated learning can exploit decentralized data for model learning and significantly reduce privacy and security risks. 
However, since model updates are communicated between the server and user clients for many rounds, the communication cost would be huge if the model is large. 
To remedy this issue, we propose a communication efficient federated learning method with knowledge distillation, which can reduce the parameters to be communicated and meanwhile keep competitive model performance.

\begin{figure*}[!t]
  \centering 
      \includegraphics[width=0.9\linewidth]{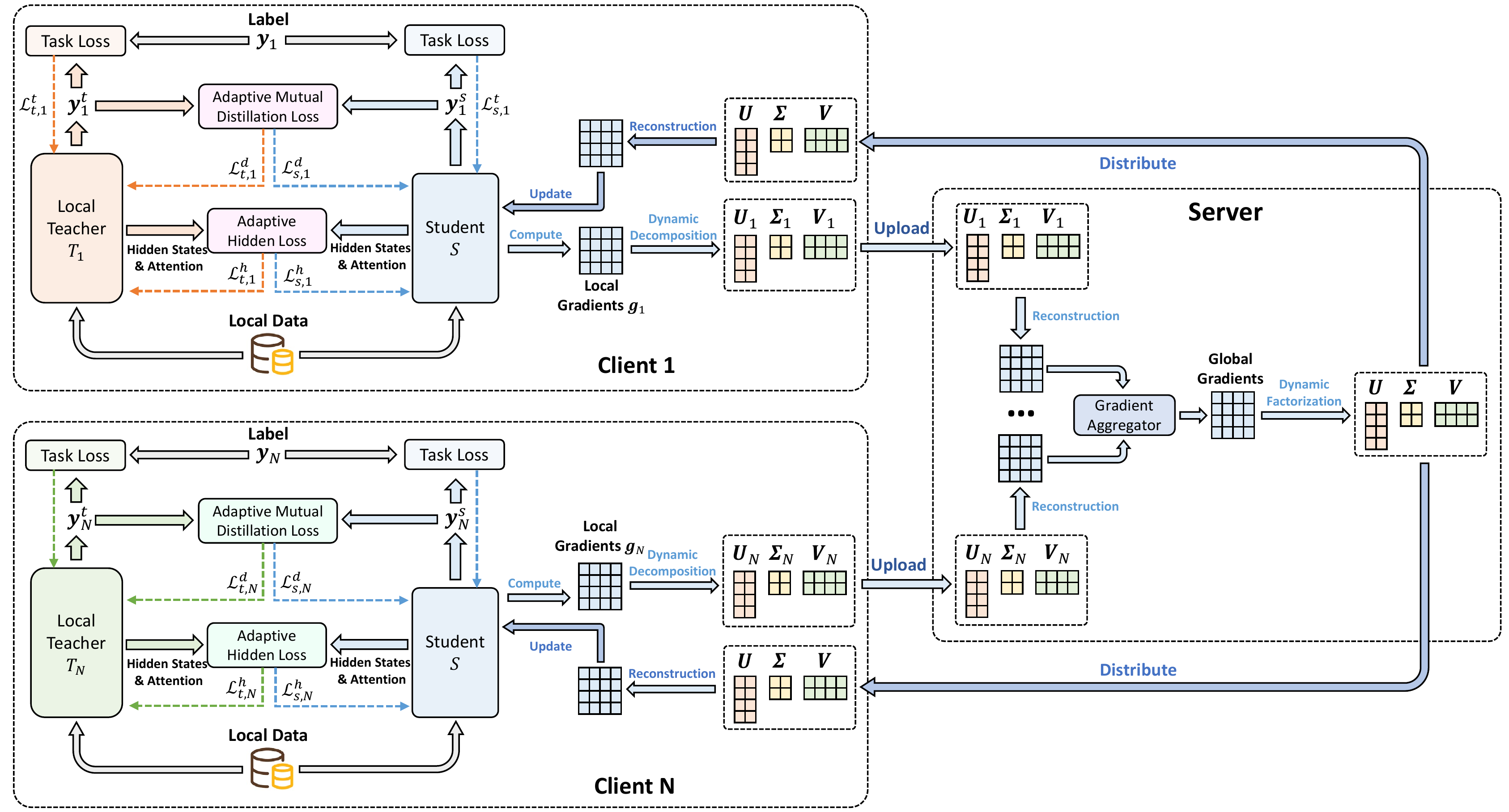}
  \caption{The framework of our \textit{FedKD} approach.}\label{fig.model}
\end{figure*}

\subsection{Knowledge Distillation}

Knowledge distillation is a technique to transfer knowledge from a large teacher model (e.g., BERT) to a small student model~\cite{hinton2015distilling}, which is widely used for model compression~\cite{sanh2019distilbert,sun2019patient,jiao2020tinybert,wang2020minilm}.
For example, ~\citet{sanh2019distilbert} proposed a DistilBERT approach that distills useful knowledge from the output using the distillation loss and the hidden states of the teacher model via a cosine loss.
 ~\citet{sun2019patient} proposed a BERT-PKD approach that aligns the hidden states of the student model with the teacher using a mean squared error loss.
~\citet{jiao2020tinybert} proposed a TinyBERT approach that can additionally transfer useful knowledge from the attention matrix of the teacher model.
However, these methods usually require centralized data storage, which may pose privacy issues during data collection.

\subsection{Communication Efficient FL}

Generally, the communication efficiency of federated learning can be improved by gradient compression~\cite{konevcny2016federated,caldas2018expanding,rothchild2020fetchsgd} and knowledge distillation~\cite{sui2020feded}.
Both genres of methods are orthogonal in reducing the communication cost of federated learning and are usually compatible with each other.
A core technique used by existing knowledge distillation-based federated learning methods is codistillation~\cite{anil2018large}.
In this method, the models on different clients are learned on the same dataset.
The output of each model is regularized to be similar to the ensemble of predictions from all models via a distillation loss.
The idea of codistillation is used by several methods to reduce communication cost of federated learning~\cite{sui2020feded,li2019fedmd,seo2020federated,lin2020ensemble,sun2020federated}.
For example,~\citet{sui2020feded} proposed a federated ensemble distillation approach for medical relation extraction.
It first learns student models locally on each client and then uses the student models to generate predictions on a shared dataset and upload them to a server.
The server ensembles the predictions from different clients as a virtual teacher and computes the distillation loss between the teacher and students.
In this way, the model parameters do not need to be uploaded, and only the predictions on the shared dataset are communicated, which can reduce the communication cost.
However, these methods require a dataset that is shared among different clients to conduct ensemble distillation.
Unfortunately, in many scenarios such as personalized recommendation, the data (e.g., user behavior logs) is highly privacy-sensitive and cannot be shared or exchange among different clients.
Thus, these methods cannot be applied to these scenarios.
By contrast, our approach circumvents the need of a shared dataset because the teacher models in our approach are locally stored on different clients.
Our approach can also effectively reduce the communication cost by communicating a distilled tiny student model instead of the original large model and using SVD to reduce gradient size.

\section{FedKD}\label{sec:Model}

In this section, we introduce our communication efficient federated learning approach based on knowledge distillation (FedKD).
We first present a definition of the problem studied in this paper, then introduce the details of our approach, and finally present some discussions on the computation and communication complexity of our approach.

\subsection{Problem Definition}

In our approach, we assume that there are $N$ clients that locally store their private data, where the raw data never leaves the client where it is stored.
We denote the dataset on the $i$-th client as $D_i$.
In our approach, each client keeps a large local teacher model  $T_i$ with a parameter set $\Theta^t_i$ and a local copy of a smaller shared student model $S$ with a parameter set $\Theta^s$.
In addition, a central server coordinates these clients for collaborative model learning.
The goal is to learn a strong model in a privacy-preserving way with less communication cost.

\subsection{Federated Knowledge Distillation}

Next, we introduce the details of our federated knowledge distillation framework, as shown in Figure~\ref{fig.model}.
In each iteration, each client simultaneously computes the update of the local teacher model and the student model based on the supervision of the labeled local data and the knowledge distilled from each other.
The teacher models are locally updated, while the student model is shared among different clients and are learned collaboratively.
Since the local teacher models have more sophisticated architectures than the student model, the useful knowledge encoded by the teacher model can help teach the student model.
In addition, since the teacher model can only learn from local data while the student model can see the data on all clients, the teacher can also benefit from the knowledge distilled from the student model.

In our approach, we use three loss functions to learn student and teacher models locally, including an adaptive mutual distillation loss to transfer knowledge from output soft labels, an adaptive hidden loss to distill knowledge from the hidden states and self-attention heatmaps, and a task loss to directly provide task-specific supervision for learning the teacher and student models.
We denote the  soft probabilities of a sample $x_i$ predicted by the local teacher and student on the $i$-th client as $\mathbf{y}^t_i$ and $\mathbf{y}^s_i$, respectively.
Since incorrect predictions from the teacher/student model may mislead the other one in the knowledge transfer, we propose an adaptive method to weight the distillation loss according to the quality of predicted soft labels.
We first use the task labels to compute the task losses for the teacher and student models (denoted as $\mathcal{L}^t_{t,i}$ and $\mathcal{L}^s_{s,i}$).
We denote the gold label of $x_i$ as $\mathbf{y}_i$, and the task losses are formulated as follows:
\begin{equation}
    \mathcal{L}^t_{t,i}=\rm{CE}(\mathbf{y}_i,\mathbf{y}^t_i),
\end{equation}
\begin{equation}
    \mathcal{L}^t_{s,i}=\rm{CE}(\mathbf{y}_i,\mathbf{y}^s_i),
\end{equation}
where $\rm{CE}$ stands for cross-entropy.
The adaptive distillation losses for both teacher and student models (denoted as $\mathcal{L}^d_{t,i}$ and $\mathcal{L}^d_{s,i}$)are formulated as follows:
\begin{equation}
   \mathcal{L}^d_{t,i}=\frac{\rm{KL}(\mathbf{y}^s_i,\mathbf{y}^t_i)}{\mathcal{L}^t_{t,i}+\mathcal{L}^t_{s,i}},
\end{equation}
\begin{equation}
\mathcal{L}^d_{s,i}=\frac{\rm{KL}(\mathbf{y}^t_i,\mathbf{y}^s_i)}{\mathcal{L}^t_{t,i}+\mathcal{L}^t_{s,i}},
\end{equation}
where $\rm{KL}$ means the Kullback–Leibler divergence.
In this way, the distillation intensity is weak if the predictions of teacher and student are not reliable.
The distillation loss becomes dominant if the student and teacher are well tuned, which has the potential to mitigate the risk of overfitting.
In addition, previous works have validated that transferring knowledge between the hidden states~\cite{sun2019patient} and hidden attention matrices~\cite{jiao2020tinybert} (if available) is beneficial for student teaching.
Thus, taking language model distillation as an example, we also introduce additional adaptive hidden losses  to align the hidden states and attention heatmaps of the student and the local teachers.
The losses for the teacher and student models (denoted as $\mathcal{L}^h_{t,i}$ and $\mathcal{L}^h_{s,i}$) are formulated as follows:
\begin{equation}
    \mathcal{L}^h_{t,i}=\mathcal{L}^h_{s,i}=\frac{\rm{MSE}(\mathbf{H}^t_i,\mathbf{W}^h_i\mathbf{H}^s)+\rm{MSE}(\mathbf{A}^t_i,\mathbf{A}^s)}{\mathcal{L}^t_{t,i}+\mathcal{L}^t_{s,i}},
\end{equation}
where $\rm{MSE}$ stands for the mean squared error, $\mathbf{H}^t_i$, $\mathbf{A}^t_i$, $\mathbf{H}^s$, and $\mathbf{A}^s$ respectively denote the hidden states and attention heatmaps in the $i$-th local teacher and the student, and $\mathbf{W}^h_i$ is a learnable linear transformation matrix.
Here we propose to control the intensity of the adaptive hidden loss based on the prediction correctness of the student and teacher. 
Besides, motivated by the task-specific distillation framework in~\cite{tang2019distilling}, we also learn the student model based on the task-specific labels on each client.
Thus, on each client the unified loss functions for computing the local update of teacher and student models (denoted as $\mathcal{L}_{t,i}$ and $\mathcal{L}_{s,i}$) are formulated as follows:
\begin{equation}
    \mathcal{L}_{t,i}=\mathcal{L}^d_{t,i}+\mathcal{L}^{h}_{t,i}+\mathcal{L}^t_{t,i},
\end{equation}
\begin{equation}
    \mathcal{L}_{s,i}=\mathcal{L}^d_{s,i}+\mathcal{L}^{h}_{s,i}+\mathcal{L}^t_{s,i},
\end{equation}
The student model gradients $\mathbf{g}_i$ on the $i$-th client can be derived  from $\mathcal{L}_{s,i}$ via $\mathbf{g}_i=\frac{\partial \mathcal{L}_{s,i}}{\partial \Theta^s}$, where $\Theta^s$ is the parameter set of student model.
The local teacher model on each client is immediately updated by their local gradients derived from the loss function $\mathcal{L}_{t,i}$.

Afterwards, the local gradients $\mathbf{g}_i$ on each client will be uploaded to the central server for global aggregation.
Since the raw model gradients may still contain some private information~\cite{zhu2020deep}, we encrypt the local gradients  before uploading.
The server receives the local student model gradients from different clients and uses a gradient aggregator\footnote{We use the FedAvg method for simplicity.} to synthesize the local gradients into a global one (denoted as $\mathbf{g}$).
The server further delivers the aggregated global gradients to each client for a local update.
The client decrypts the global gradients to update its local copy of the student model.
This process will be repeated until both student and teacher models converge.
Note that in the test phase, the teacher model is used for label inference.

\subsection{Dynamic Gradients Approximation}

In our \textit{FedKD} framework, although the size of student model updates is smaller than the teacher models, the communication cost can still be relatively high when the student model is not tiny.
Thus, we propose to a dynamic gradients approximation method to compress the gradients exchanged among the server and clients to further reduce computational cost.
As shown in Fig.~\ref{fig.model}, we first factorize the local gradients into smaller matrices before uploading them. 
The server reconstructs the local gradients by multiplying the factorized matrices before aggregation.
The aggregated global gradients are further factorized, which are distributed to the clients for reconstruction and model update.
More specifically, we denote the gradient $\mathbf{g}_i\in \mathbb{R}^{P\times Q}$ as a matrix with $P$ rows and $Q$ columns (we assume $P\geq Q$).\footnote{We formulate $\mathbf{g}_i$ as a single matrix for simplicity. In practice, different parameter matrices in the model are factorized independently. The global gradients on the server are factorized in the same way.}
It is approximately factorized into the multiplication of three matrix, i.e., $\mathbf{g}_i \approx \mathbf{U}_i\mathbf{\Sigma}_i\mathbf{V}_i$, where $\mathbf{U}_i\in \mathbb{R}^{P\times K}$, $\mathbf{\Sigma}_i\in \mathbb{R}^{K\times K}$, 
 $\mathbf{V}_i\in \mathbb{R}^{K\times Q}$ are factorized matrices and $K$ is the number of retained singular values.
If the value of $K$ satisfies $PK+K^2+KQ<PQ$, the size of uploaded and downloaded gradients can be reduced.
We denote the singular values of $\mathbf{g}_i$ as $[\sigma_1, \sigma_2, ..., \sigma_Q]$ (ordered by their absolute values).
To control the approximation error, we use an energy threshold $T$ to decide how many singular values are kept, which is formulated as follows:
\begin{equation}
    \min_K \frac{\sum_{i=1}^K \sigma_i^2}{\sum_{i=1}^Q \sigma_i^2}>T.
\end{equation}
To better help the model converge,  we propose to use a dynamic value of $T$.
The function between the threshold $T$ and the percentage of training steps $t$ is formulated as follows:
\begin{equation}
    T(t)=T_{start}+(T_{end}-T_{start})t, t\in [0, 1],
\end{equation}
where $T_{start}$ and $T_{end}$ are two hyperparameters that control the start and end values of $T$.
In this way, the student model is learned on roughly approximated gradients at the beginning, while learned on more accurately approximated gradients when the model gets to convergence, which can help learn a more accurate student model.

To help readers better understand how \textit{FedKD} works,  we summarize the entire workflow of \textit{FedKD} in the Algorithm 1 in Appendix. 

\subsection{Complexity Analysis}

In this section, we will present some analysis on the complexity of our \textit{FedKD} approach in terms of computation and communication cost.
We denote the number of communication rounds as $R$ and the average size of dataset on each client as $D$.
Thus, the computational cost of directly learning a large model (the parameter set is denoted as $\Theta^t$) in a federated way is $O(RD|\Theta^t|)$, and the communication cost is $O(R|\Theta^t|)$.\footnote{We assume the cost is linearly proportional to model sizes.}
In \textit{FedKD}, the communication cost is $O(R|\Theta^s|/\rho)$ ($\rho$ is the gradient compression ratio), which is much smaller because $|\Theta^s|\ll |\Theta^t|$ and $\rho>1$.
The computational cost contains three parts, i.e., local teacher model learning, student model learning and gradient compression/reconstruction, which are $O(RD|\Theta^t|)$, $O(RD|\Theta^s|)$ and $O(RPQ^2)$, respectively.
The total computational cost of \textit{FedKD} is $O(RD|\Theta^t|+RD|\Theta^s|+RPQ^2)$.
In practice, compared with the standard FedAvg~\cite{mcmahan2017communication} method, the extra computational cost of learning the student model in \textit{FedKD} is much smaller than learning the large teacher model, and SVD can also be very efficiently computed in parallel.
Thus, \textit{FedKD} is efficient in terms of both communication and computation.

\section{Experiments}\label{sec:Experiments}

\subsection{Datasets and Experimental Settings}

Our experiments are conducted in two tasks that involve user data.
The first one is personalized news recommendation, which needs to predict whether a user will click a candidate news based on the user interest inferred from historical news click behaviors.
In this task we use the \textit{MIND}~\cite{wu2020mind} dataset.\footnote{https://msnews.github.io/}
It contains the news impression logs of 1 million users on the Microsoft News platform during 6 weeks.
The logs in the last week are used for test, and the rest are for training and validation.
The second one is adverse drug reaction (ADR) mentioning tweet detection, which is a binary classification task.
We use the dataset released by the 3rd shared task of the SMM4H 2018 workshop~\cite{weissenbacher2018overview}.\footnote{https://healthlanguageprocessing.org/smm4h18}
We denote this dataset as \textit{SMM4H}.
The original \textit{SMM4H} dataset contains 25,678 tweet IDs.
However, since many tweet texts in this dataset are no longer available, we only crawled 16,694 tweets for experiments.
Following~\cite{wu2019msa}, we use 80\% of the dataset for training, 10\% for validation and 10\% for test.
The detailed statistics of these two datasets are summarized in Table~\ref{dataset}.
To simulate the scenario where private data is decentralized on different clients, we randomly divide the training data into 4 folds and assume that each fold is locally stored on different clients. 

\begin{table}[h]
\centering
\resizebox{0.46\textwidth}{!}{ 
\begin{tabular}{lrlr}
 \Xhline{1.5pt}
\multicolumn{4}{c}{MIND}                                             \\ \hline
\# users              & 1,000,000 & \# impressions      & 15,777,377 \\
\# news               & 161,013   & \# clicks           & 24,155,470 \\
avg. title len.       & 11.52     & \# training samples & 2,186,683  \\
\# validation samples & 365,200   & \# test samples     & 2,341,619  \\ \hline
\multicolumn{4}{c}{SMM4H}                                            \\ \hline
\# tweets             & 16,694    & \# positives        & 1,355      \\
avg. tweet len.       & 16.48     & \# negatives        & 15,336     \\ \Xhline{1.5pt}
\end{tabular}
}

\caption{Statistics of the datasets.}\label{dataset}
\end{table}

In our experiments, on each client we use the UniLM-Base~\cite{bao2020unilmv2} model as the local teacher.\footnote{We take pre-trained language model distillation as a representative example in our experiments.}
We use its submodels with the first 4 or 2 Transformer layers as the student models. 
On the \textit{MIND} dataset, we incorporate the language model as the news encoder of \textit{NAML}. On the \textit{SMM4H} dataset we apply an attentive pooling and a dense layer after the language model for text classification.
The energy thresholds $T_{start}$ and $T_{end}$ are 0.95 and 0.98, respectively.
The optimizer we use is Adam~\cite{kingma2014adam}.
\footnote{The detailed hyperparameter settings of our approach and baselines are in the Appendix.}
Following~\cite{wu2020mind}, on the \textit{MIND} dataset, we use AUC, MRR, nDCG@5 and nDCG@10 as the metrics.
On the \textit{SMM4H} dataset, we use precision, recall and Fscore of the positive class as the metrics~\cite{wu2019msa}.
We repeat each experiment repeat 5 times to mitigate occasionality.

\subsection{Performance Evaluation}

\begin{table}[t]
\resizebox{0.48\textwidth}{!}{
\begin{tabular}{lccccc}
 \Xhline{1.5pt}
\textbf{Methods} & \textbf{AUC} & \textbf{MRR} & \textbf{nDCG@5} & \textbf{nDCG@10} & \textbf{\begin{tabular}[c]{@{}c@{}} Comm. Cost\\ per Client\end{tabular}} \\ \hline
UniLM (Local)            & 68.8$\pm$0.5  & 33.5$\pm$0.4  & 36.6$\pm$0.5  & 42.4$\pm$0.6  & -      \\
UniLM (Cen)          & \textbf{71.0}$\pm$0.1  & \textbf{35.8}$\pm$0.1  & \textbf{39.0}$\pm$0.1  & \textbf{44.8}$\pm$0.1  & -  \\
UniLM (Fed)          & 70.9$\pm$0.3  & 35.7$\pm$0.2  & 38.9$\pm$0.3  & 44.7$\pm$0.4  & 2.05GB  \\ \hline
DistilBERT$_6$      & 69.3$\pm$0.2  & 34.0$\pm$0.2  & 37.5$\pm$0.2  & 43.0$\pm$0.1  &  1.03GB  \\
DistilBERT$_4$      & 69.0$\pm$0.2  & 33.7$\pm$0.1  & 37.0$\pm$0.1  & 42.6$\pm$0.2  &  0.69GB  \\
BERT-PKD$_6$        & 69.6$\pm$0.2  & 34.4$\pm$0.3  & 37.7$\pm$0.3  & 43.4$\pm$0.2  &  1.03GB  \\
BERT-PKD$_4$        & 69.2$\pm$0.2  & 33.8$\pm$0.2  & 37.1$\pm$0.3  & 42.9$\pm$0.3  &  0.69GB  \\
TinyBERT$_6$        & 69.7$\pm$0.2  & 34.5$\pm$0.2  & 37.9$\pm$0.1  & 43.5$\pm$0.2  &  1.03GB  \\
TinyBERT$_4$        & 69.4$\pm$0.3  & 33.9$\pm$0.3  & 37.5$\pm$0.2  & 43.1$\pm$0.2  &  0.17GB  \\ \hline
UniLM$_4$           & 69.6$\pm$0.1  & 34.4$\pm$0.2  & 37.7$\pm$0.1  & 43.4$\pm$0.2  &  0.69GB  \\
UniLM$_2$           & 68.9$\pm$0.2  & 33.6$\pm$0.2  & 36.8$\pm$0.2  & 42.5$\pm$0.1  &  0.35GB  \\ \hline
FetchSGD                               & 70.5$\pm$0.4  & 35.2$\pm$0.3  & 38.2$\pm$0.3  & 44.0$\pm$0.4  &  0.51GB  \\
FedDropout                               & 70.5$\pm$0.2  & 35.1$\pm$0.2  & 38.3$\pm$0.3  & 44.2$\pm$0.3  &  1.23GB  \\ \hline
FedKD$_4$                               & \textbf{71.0}$\pm$0.1  & 35.6$\pm$0.1  & 38.9$\pm$0.1  & \textbf{44.8}$\pm$0.1  &  0.19GB  \\
FedKD$_2$                               & 70.5$\pm$0.1  & 35.3$\pm$0.2  & 38.6$\pm$0.1  & 44.3$\pm$0.2  &  \textbf{0.11GB}  \\
 \Xhline{1.5pt}
\end{tabular}
}
\caption{Performance of different methods on \textit{MIND}.} \label{table.performance1} 
\end{table}

\begin{table}[t]
\resizebox{0.48\textwidth}{!}{
\begin{tabular}{lcccc}
 \Xhline{1.5pt}
\textbf{Methods} & \textbf{Precision} & \textbf{Recall} & \textbf{Fscore} & \textbf{\begin{tabular}[c]{@{}c@{}} Comm. Cost\\ per Client\end{tabular}} \\ \hline
UniLM (Local)           & 53.2$\pm$1.3   & 54.6$\pm$1.4  & 53.9$\pm$1.1  & -                   \\
UniLM (Cen)           & \textbf{60.3}$\pm$0.7   & 61.6$\pm$0.8  & \textbf{60.8}$\pm$0.4  & -                   \\
UniLM (Fed)           & 59.1$\pm$0.6   & 62.3$\pm$0.6  &60.6$\pm$0.4  & 1.37GB               \\ \hline
DistilBERT$_6$     & 56.8$\pm$0.8   & 59.2$\pm$0.8  & 57.9$\pm$0.5                  &0.69GB               \\
DistilBERT$_4$     & 56.5$\pm$0.9   & 58.4$\pm$1.1  & 57.1$\pm$0.7                  & 0.46GB               \\
BERT-PKD$_6$       & 56.9$\pm$0.9   & 60.4$\pm$0.8  & 58.4$\pm$0.6                  & 0.69GB               \\
BERT-PKD$_4$       & 56.3$\pm$1.1   & 59.9$\pm$0.7  & 58.0$\pm$0.6                  & 0.46GB               \\
TinyBERT$_6$       & 57.4$\pm$0.8   & 60.5$\pm$0.6  & 58.6$\pm$0.5                  & 0.69GB              \\
TinyBERT$_4$       & 57.0$\pm$0.7   & 59.9$\pm$1.2  & 58.3$\pm$0.7                  & 0.12GB                \\ \hline
UniLM$_4$          & 56.1$\pm$0.9   & 60.6$\pm$0.9  & 58.2$\pm$0.5                  & 0.46GB               \\
UniLM$_2$          & 53.8$\pm$0.8   & 59.1$\pm$1.0  & 56.3$\pm$0.6                  & 0.24GB               \\ \hline
FetchSGD          & 57.5$\pm$0.9   & 60.4$\pm$1.1  & 59.0$\pm$0.8                  & 0.34GB               \\
FedDropout          & 57.8$\pm$1.0   & 61.0$\pm$0.8  & 59.4$\pm$0.6                  & 0.82GB               \\\hline
FedKD$_4$           & 59.4$\pm$0.6 & \textbf{62.8}$\pm$0.9  & 60.7$\pm$0.5   & 0.12GB         \\
FedKD$_2$           & 58.2$\pm$0.7 & 62.4$\pm$0.9  & 59.8$\pm$0.6    & \textbf{0.07GB} \\ \Xhline{1.5pt}
\end{tabular}
}
\caption{Performance of different methods on \textit{SMM4H}.} \label{table.performance2} 
\end{table}

First, we compare the performance and communication cost\footnote{The communication cost on the two datasets are slightly different due to the number of updated token embeddings.} of \textit{FedKD} with several additional baselines, including:
(1) \textit{UniLM (Local)}, learning the full UniLM model with the local data on a client;
(2) \textit{UniLM (Cen)}, learning the full UniLM model on centralized data;
(3) \textit{UniLM (Fed)}, learning the full UniLM model in the standard federated framework;
(4) \textit{DistilBERT}~\cite{sanh2019distilbert}, finetuning the DistilBERT model in federated learning;
(5) \textit{BERT-PKD}~\cite{sun2019patient}, finetuning BERT-PKD in a federated manner;
(6) \textit{TinyBERT}~\cite{jiao2020tinybert}, finetuning TinyBERT in a federated way;
(7) \textit{UniLM$_{4/2}$}, using the first 4 or 2 layers of UniLM in federated learning.
(8) \textit{FetchSGD}~\cite{rothchild2020fetchsgd}, a count sketch based communication efficient federated learning method.
(9) \textit{FedDropout}~\cite{caldas2018expanding}, a federated dropout method to reduce the number of exchanged parameters.
In the methods (4)-(6), we compare the performance of their officially released 6-layer and 4-layer models.
In methods (8) and (9), we use the full UniLM model.
The results on the \textit{MIND} and \textit{SMM4H} datasets are respectively shown in  Tables~\ref{table.performance1} and~\ref{table.performance2}.
From the results, we have the following findings.
First, compared with \textit{UniLM (local)}, other methods achieve better performance.
This is because the local data on a single client may not be sufficient to learn a strong model, while federated learning can exploit data decentralized on multiple clients to facilitate model training.
Second, although \textit{UniLM} achieves the best performance, the communication cost for model learning is huge (e.g., over 2GB for each client on the \textit{MIND} dataset).
Thus, it may be difficult to incorporate it in real-world applications.
Third, compared with the off-the-shelf distilled models like \textit{DistilBERT}, \textit{BERT-PKD} and \textit{TinyBERT}, our \textit{FedKD} approach performs better.
This is because the former methods are distilled in a task-agnostic manner, which may be suboptimal in downstream tasks without further task-specific distillation.
Fourth, \textit{FedKD} also outperforms \textit{UniLM$_4$} and \textit{UniLM$_2$}.
This is because \textit{FedKD} can learn useful knowledge from the output and intermediate results of the sophisticated local teacher models while \textit{UniLM$_4$} and \textit{UniLM$_2$} cannot.
Fifth,  \textit{FedKD} can achieve better performance and lower communication cost than other communication efficient methods like \textit{FedtchSGD} and \textit{FedDropout}.
It is because \textit{FedKD} can transfer  rich knowledge between the teacher and student models to improve the model performance, and can reduce the communication cost by exchanging the updates of a small student model and meanwhile compress the gradients with SVD. 
Sixth, the communication cost of \textit{FedKD} is much less than the original \textit{UniLM} model, and the performance of \textit{FedKD} is comparable with \textit{UniLM (Fed)} and \textit{UniLM (Cen)}.
These results show that \textit{FedKD} can effectively reduce the communication cost of federated learning while keeping good performance.

\begin{figure}[!t]
  \centering
    \includegraphics[width=0.36\textwidth]{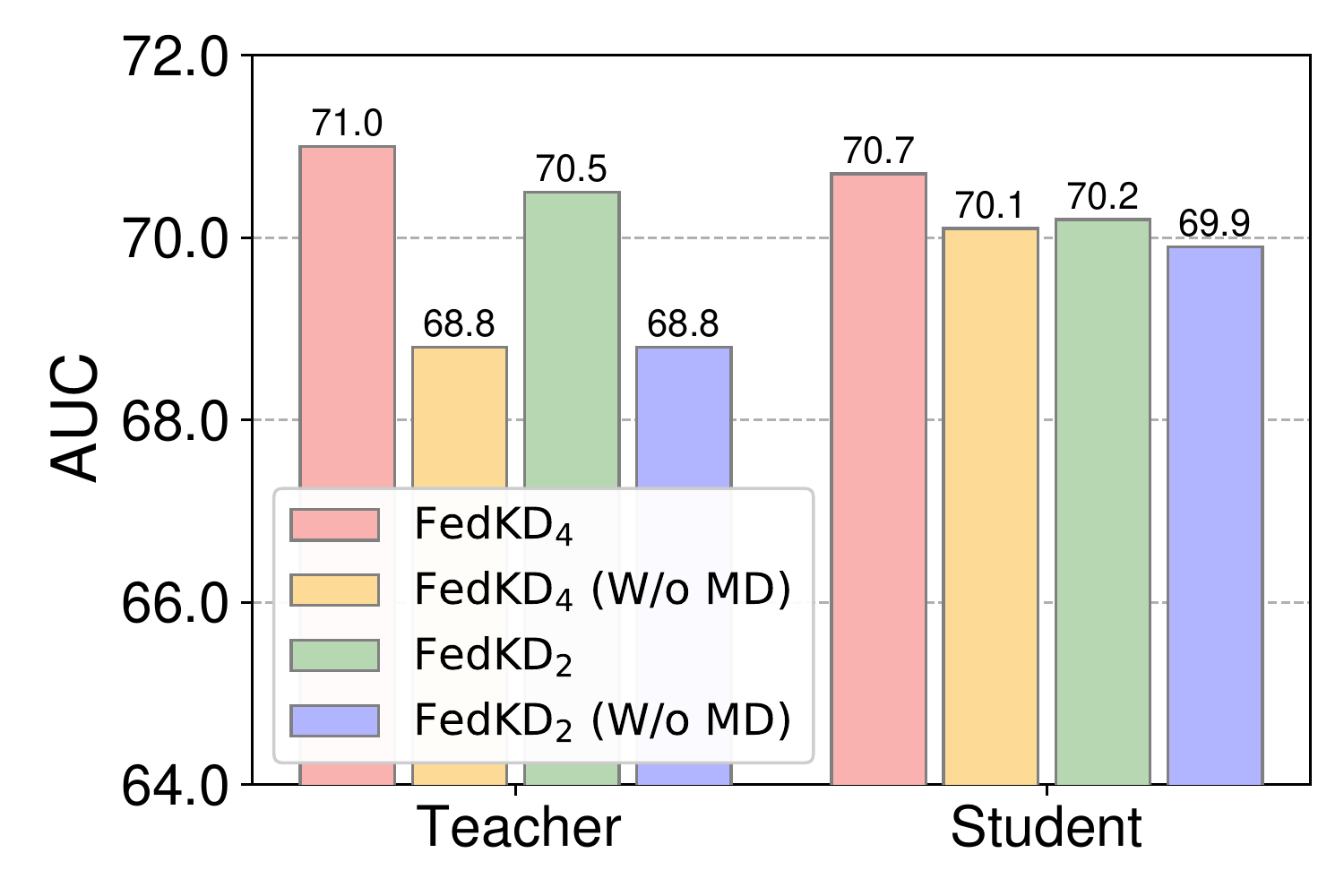}
  \caption{Influence of mutual distillation on the student and teacher models. }\label{fig.mutual1}

\end{figure}

\begin{figure}[!t]
  \centering
    \includegraphics[width=0.36\textwidth]{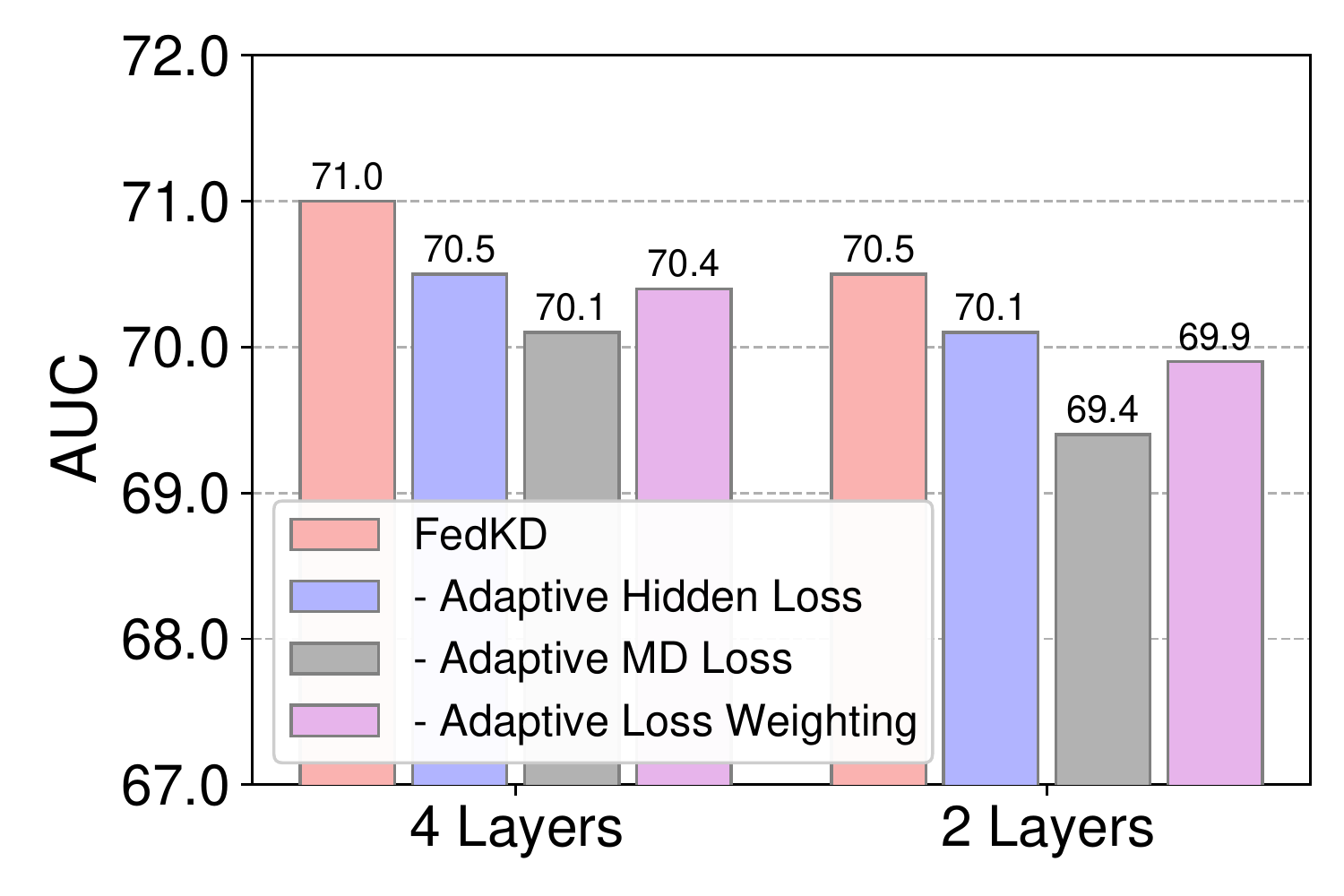}

  \caption{Effect of adaptive mutual distillation. }\label{fig.ab1}

\end{figure}

\subsection{Effectiveness of Adaptive Mutual Distillation}

We also verify the effectiveness of our proposed adaptive mutual distillation method.
We first compare the performance of \textit{FedKD} models trained with or without mutual distillation (the teacher model is only learned on local data), as shown in Fig.~\ref{fig.mutual1}.\footnote{We only include results on \textit{MIND} due to space limit. The results on \textit{SMM4H} are in Appendix.}
We observe that mutual distillation can effectively improve the performance of both teacher and student models with different sizes, especially the teacher model.
This is because useful knowledge transferred between the teachers and student can help student better  imitate the complicated teacher models, and can help teachers break the limitation of the amount of local labeled data.
In addition, we observe that local teachers slightly outperform the student.
Thus, we choose to use the teacher models for inference in the test stages.

We further compare \textit{FedKD} and its variants with the adaptive mutual distillation loss, the adaptive hidden loss or the adaptive loss weighting method removed, as shown in Fig.~\ref{fig.ab1} (we report the performance of teacher models).
We can see that both adaptive mutual distillation and adaptive hidden losses are useful for improving the model performance.
In addition, the performance is suboptimal when the adaptive loss weighting method is removed (this variant is similar to the standard mutual distillation~\cite{zhang2018deep}).
This is because weighting the distillation and hidden losses can be aware of the correctness of model predictions, which may help distill higher-quality knowledge and meanwhile mitigate the risk of overfitting.

\begin{figure}[!t]
  \centering
    \includegraphics[width=0.34\textwidth]{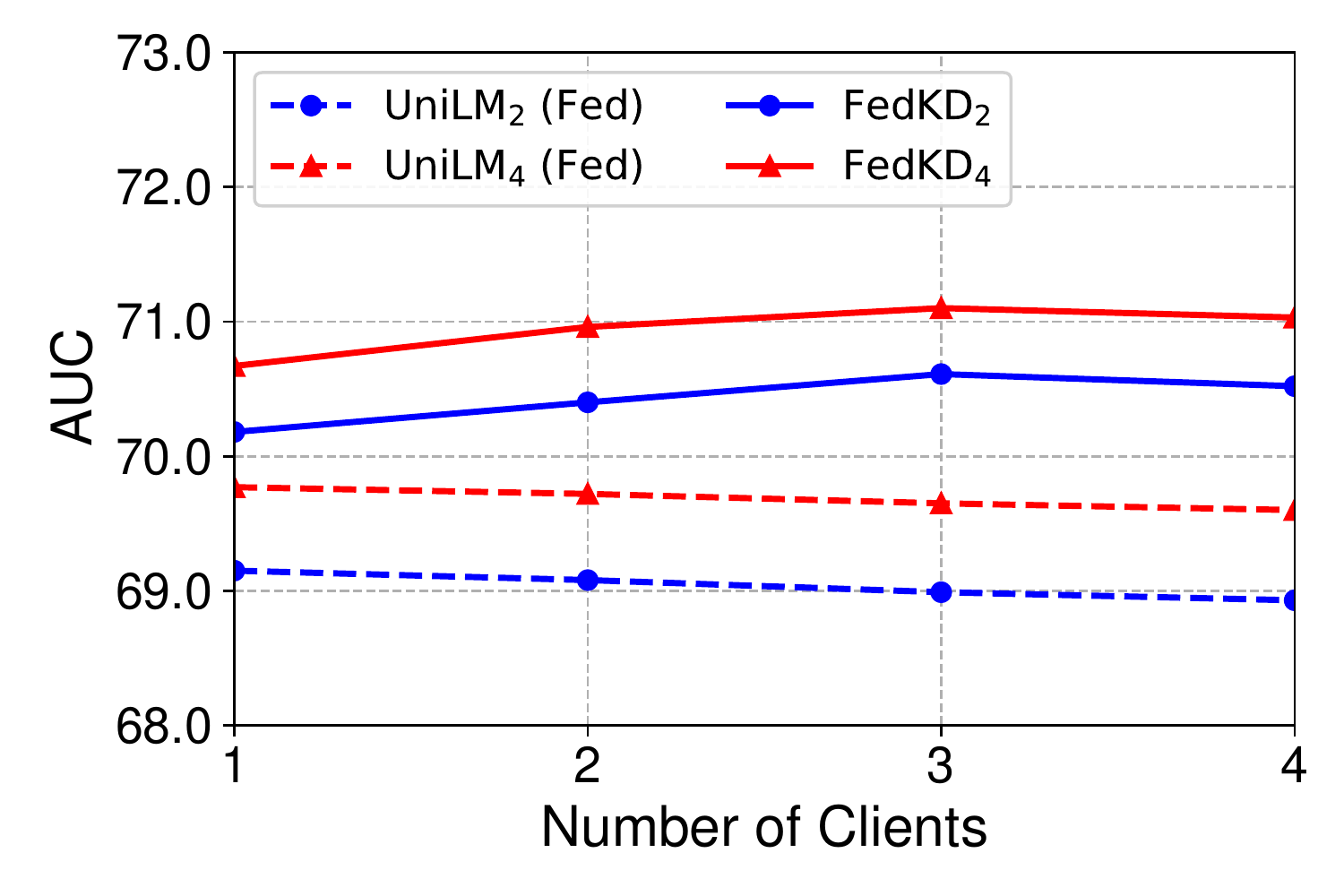}
  \caption{Influence of client number.}\label{fig.plat}
\end{figure}

\subsection{Influence of Client Number}

We study the influence of client number on the model performance in this section.
We divide the full training data into different numbers of folds to simulate the scenarios with different amounts of labeled data on each client.
Figure~\ref{fig.plat} shows the performance of \textit{FedKD} and \textit{UniLM$_{4/2}$} under different numbers of clients.
We find the performance of \textit{FedKD} is similar and can even be slightly improved when more clients are involved.
This is because \textit{FedKD} can learn from multiple teacher models on different clients, which can encode richer knowledge when more teacher models participate.
On the contrary, the performance of \textit{UniLM$_{4/2}$ (Fed)} slightly declines with the increase of client number.
This may be because the vanilla FedAvg method has some performance sacrifice by learning models for multiple epochs on limited local data.

\begin{figure}[!t]
  \centering
  
  \subfigure[$T_{start}$.]{
    \includegraphics[width=0.223\textwidth]{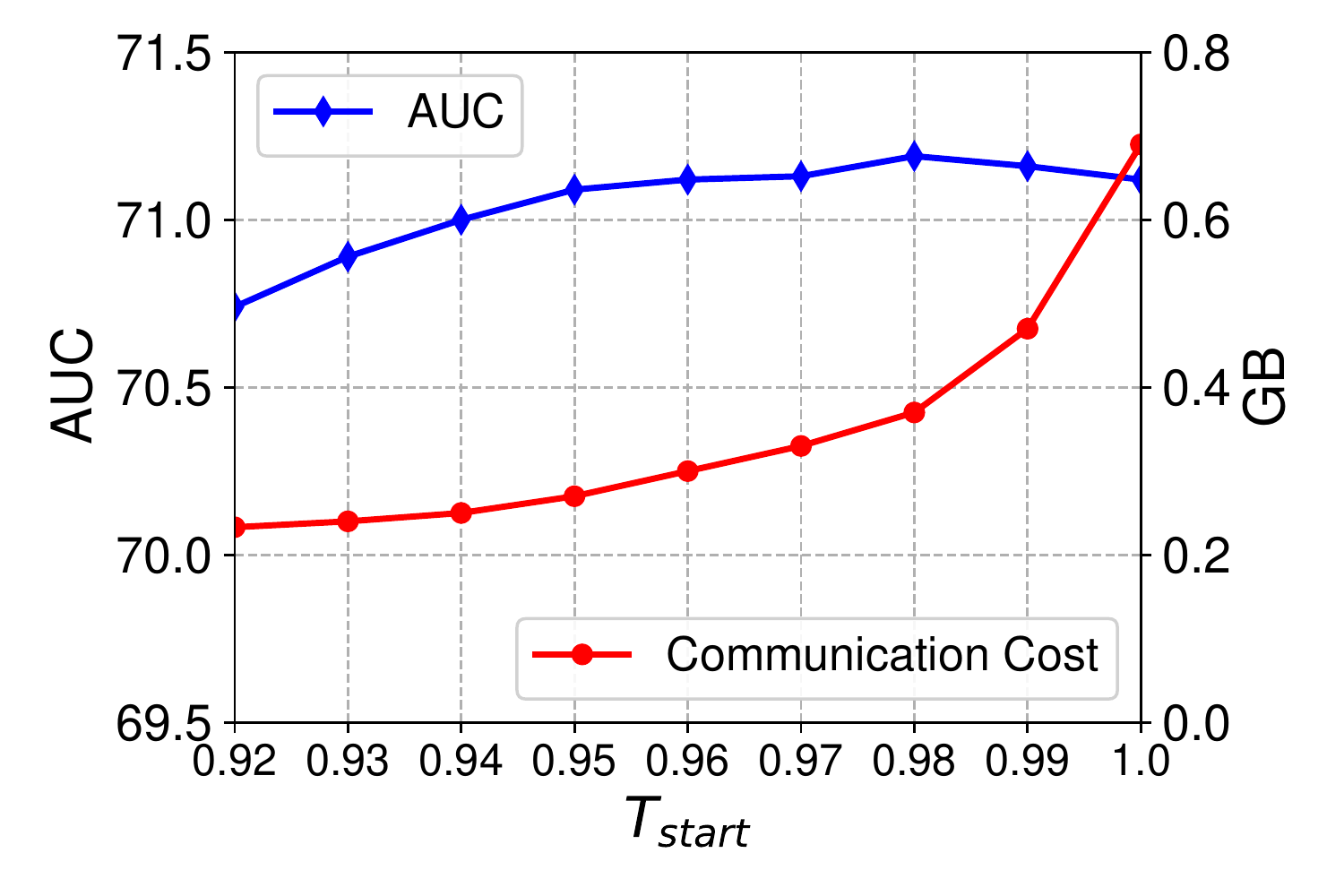}\label{fig.com1}
    }
      \subfigure[$T_{end}$.]{
    \includegraphics[width=0.223\textwidth]{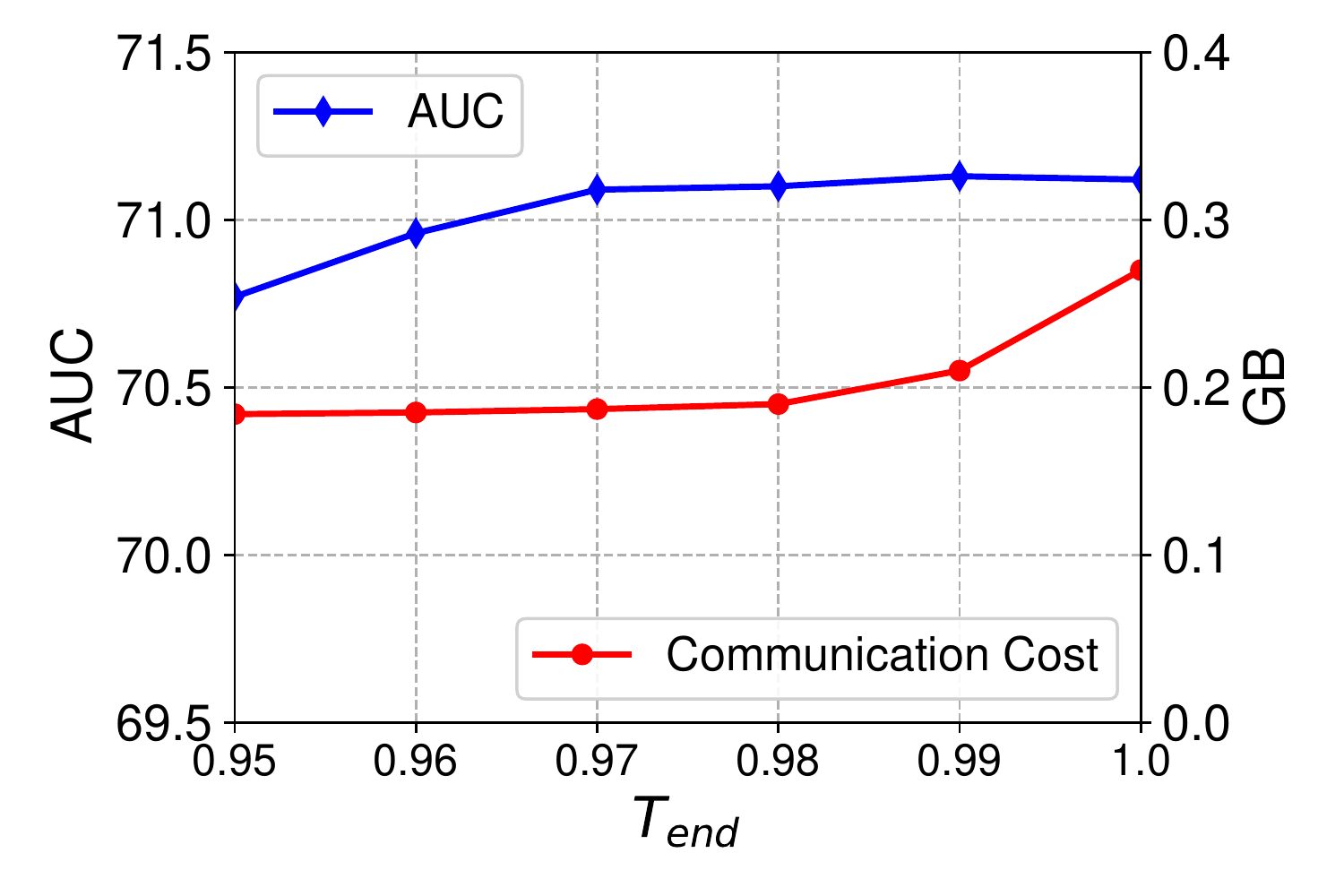}\label{fig.com3}
    }
  \caption{Influence of $T_{start}$ and $T_{end}$ on model performance and communication cost. }

\end{figure}

\begin{figure}[!t]
  \centering
  \subfigure[Beginning of training.]{
    \includegraphics[width=0.223\textwidth]{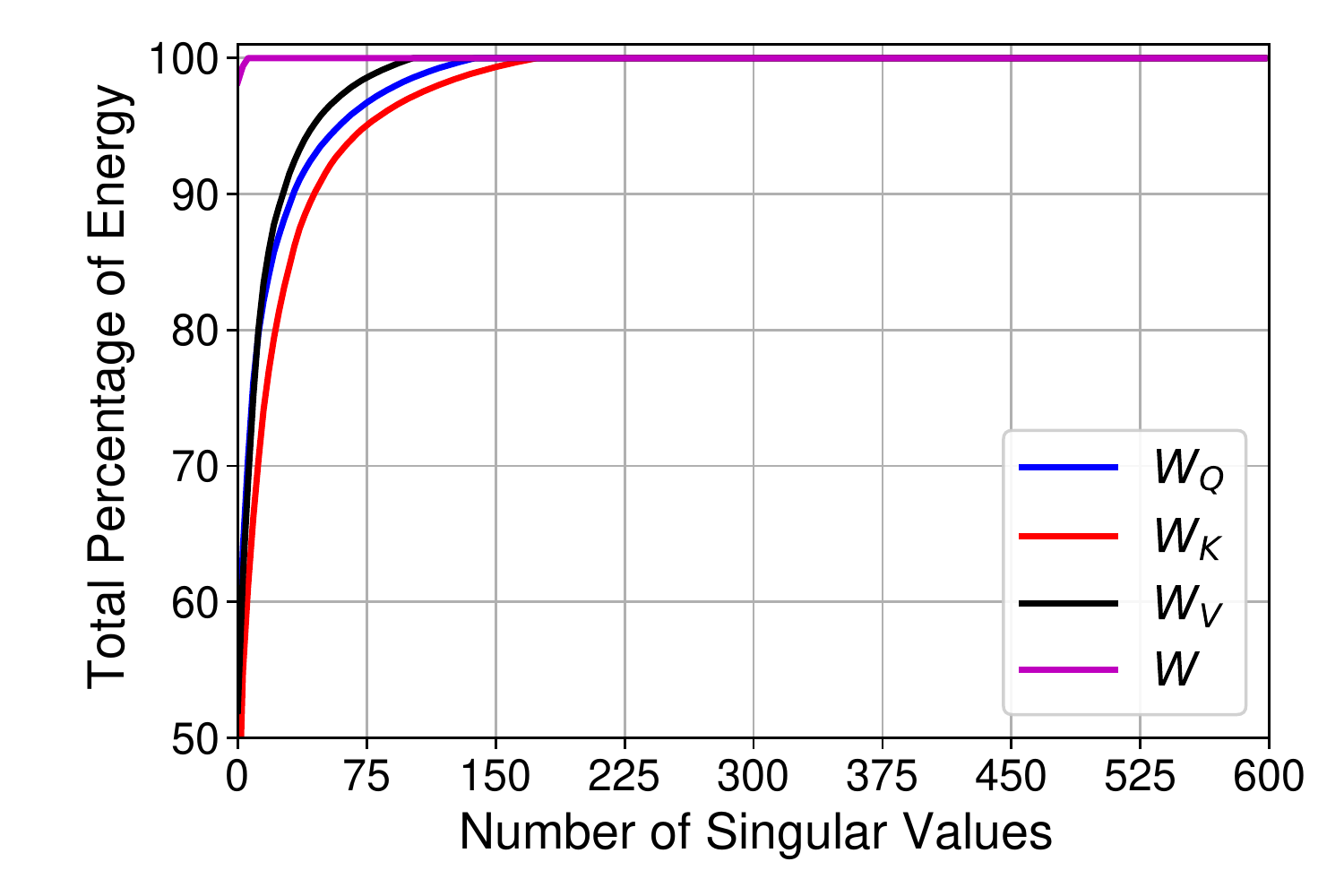}
  \label{fig.v1}
  }
   \subfigure[End of training.]{
      \includegraphics[width=0.223\textwidth]{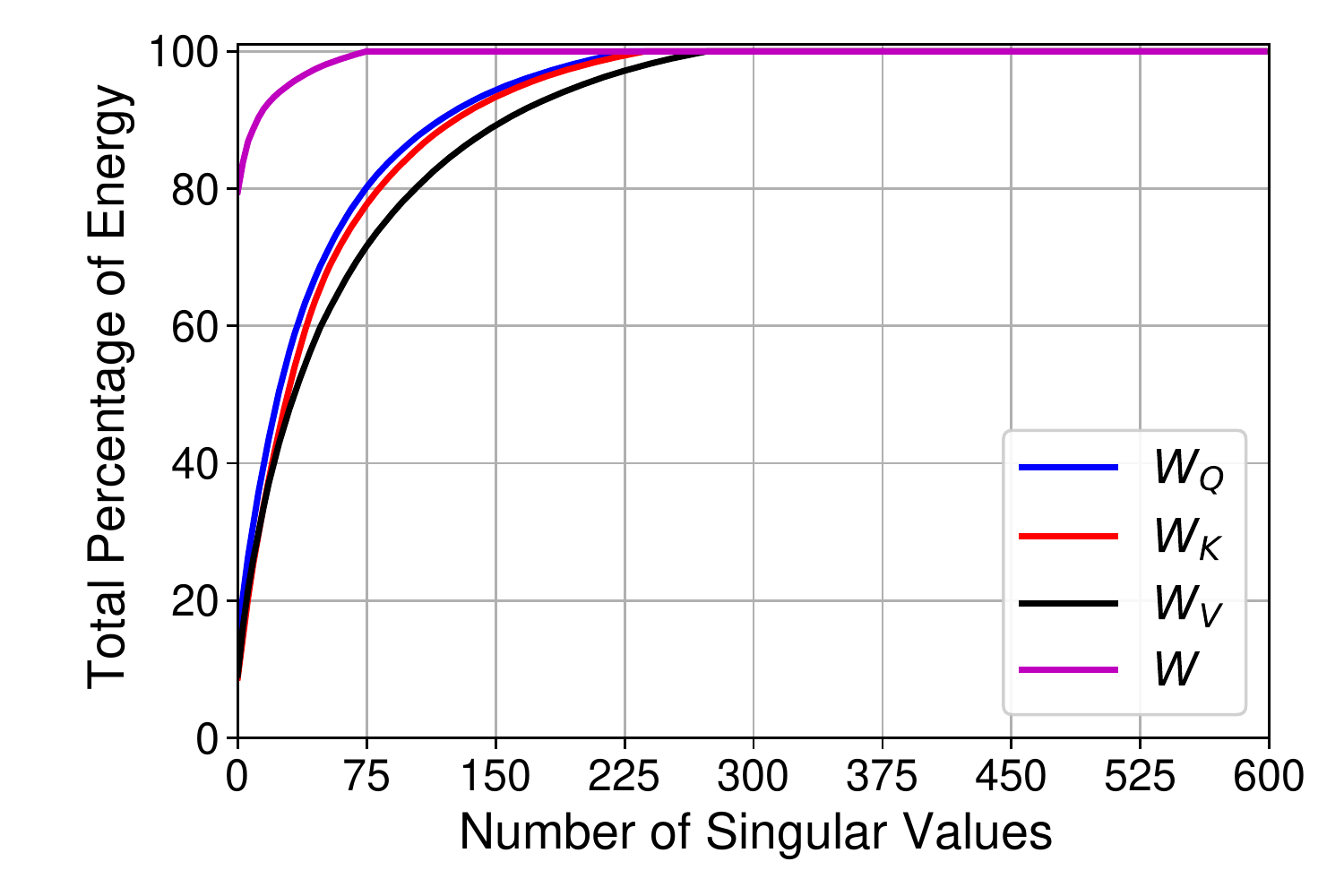}  \label{fig.v2}
      
  }
  \caption{Cumulative energy distributions of singular values of different parameter gradient matrices. $W_Q$: query parameters, $W_K$: key parameters, $W_V$: value parameters, $W$: feed-forward network parameters. }\label{fig.v}

\end{figure}

\subsection{Impact of Energy Threshold}

We then study the influence of the energy threshold $T_{start}$ and $T_{end}$ on the performance and communication cost of our approach.
We first vary $T_{start}$ under $T_{end}=1$, and the results are shown in Fig.~\ref{fig.v1}.
We find the communication cost is smaller when $T_{start}$ is smaller, while we observe that the performance starts to drop quickly when $T_{start}<0.95$.
Thus, we chose $T_{start}=0.95$ to balance communication cost and model performance.
Under $T_{start}=0.95$, we then vary $T_{end}$ to compare the performance and communication cost, as shown in Fig.~\ref{fig.v2}.
In a similar way, we choose $T_{end}=0.98$ to achieve a good tradeoff between model accuracy and communication cost.

\begin{figure}[!t]
  \centering
    \includegraphics[width=0.3\textwidth]{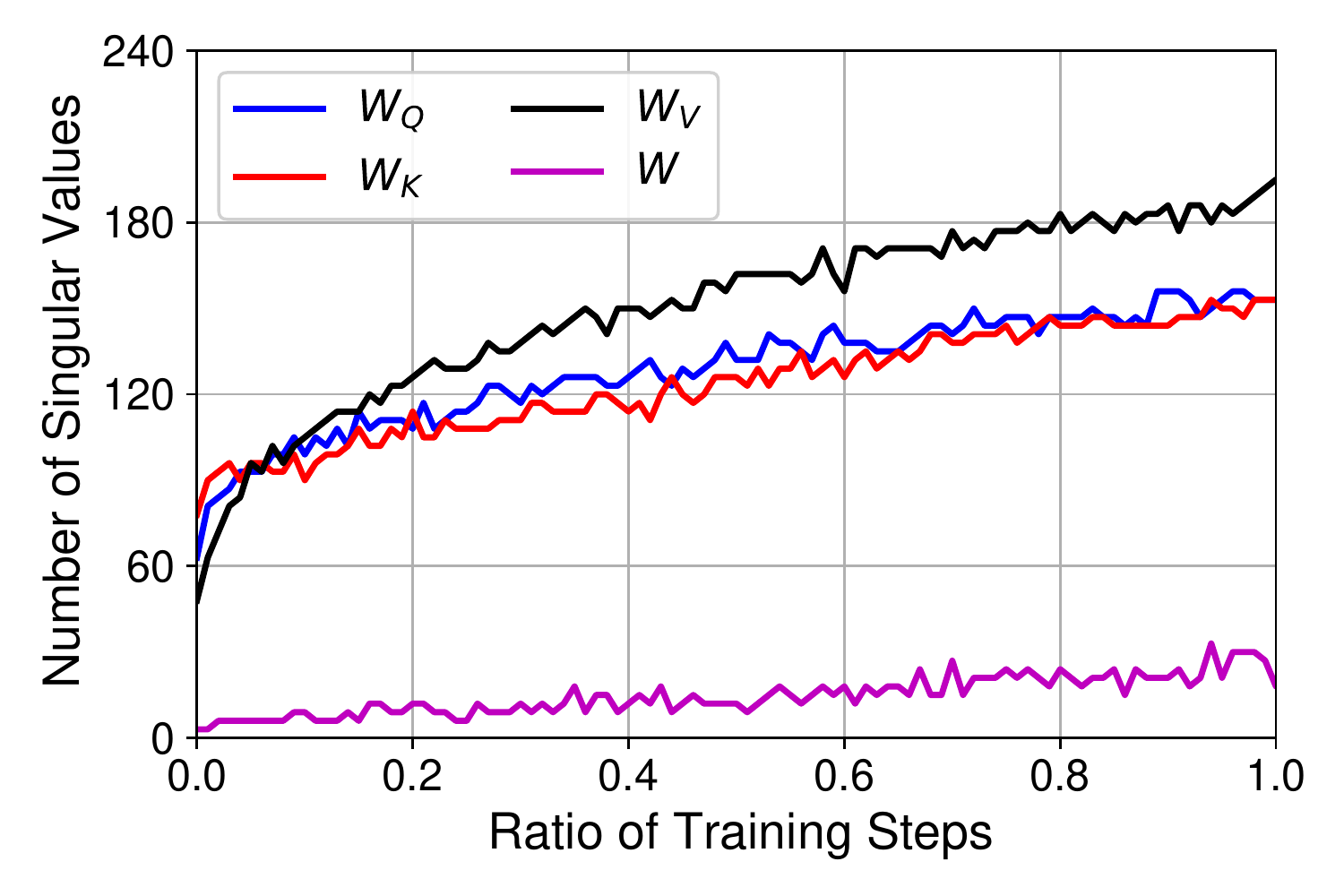}

  \caption{Evolution of the number of required singular values under $T=0.95$. }\label{fig.v3}

\end{figure}

\subsection{Analysis of Dynamic Gradient Approximation}

Finally, we present some analysis of our proposed SVD-based gradient compression method.
We show the cumulative energy distributions of singular values of different parameter gradient matrices in the UniLM model in Fig.~\ref{fig.v}, which reveals several interesting findings.
First, all kinds of parameter matrices in UniLM are low-rank, especially the parameters in the feed-forward network.
Thus, the communication cost can be greatly reduced by compressing the low-rank gradient matrices. 
In addition, we find the singular value energy is more  concentrated at the beginning than the end of training.
This may be because when the model is not well-tuned, the gradients may have more low frequency components that aim to push the model to converge more quickly.
However, when the model gets to converge, the updates of model parameters are usually subtle, which yields more high frequency components.
The evolution of required singular values under $T=0.95$ is shown in Fig.~\ref{fig.v3}.
We can see that more singular values need to be retained to achieve the same energy threshold.
To ensure the model accuracy of \textit{FedKD}, we choose to set a higher as the model training continues to learn more accurate models.

\section{Conclusion}\label{sec:Conclusion}

In this paper, we propose a communication efficient federated learning method based on knowledge distillation named \textit{FedKD}.
In our approach, we propose an adaptive mutual distillation method to reciprocally learn a teacher model and a  student model on each client, where the distillation intensity is controlled by their prediction correctness.
The large teacher model is locally updated, while the small student model is shared among different clients and learned collaboratively, which can effectively reduce the communication cost.
In addition, we propose a dynamic gradient approximation method  to further reduce the communication cost.
Extensive experiments on two benchmark datasets for different tasks validate that \textit{FedKD} can largely reduce the  communication cost in federated learning while keeping promising model performance.

\bibliography{emnlp2021}
\bibliographystyle{acl_natbib}

\clearpage
\appendix
\section{Appendix}

\subsection{Comparison with Additional Baselines}
To provide benchmarks on the \textit{MIND} and \textit{SMM4H} datasets, we compare the performance of our \textit{FedKD} approach with several baseline methods on these datasets. 
On the \textit{MIND} dataset, the additional baseline methods to be compared include: (1) \textit{EBNR}~\cite{okura2017embedding}, embedding-based news recommendation with GRU; (2) \textit{DKN}~\cite{wang2018dkn}, deep knowledge network for news recommendation; (3) \textit{NPA}~\cite{wu2019npa}, news recommendation with personalized attention; (4) \textit{NAML}~\cite{wu2019}, news recommendation with attentive multi-view learning; (5) \textit{LSTUR}~\cite{an2019neural}, news recommendation with long short-term user interest; (6) \textit{NRMS}~\cite{wu2019nrms}, news recommendation with multi-head self-attention; (7) \textit{FIM}~\cite{wang2020fine}, fine-grained interest matching for news recommendation.
On the \textit{SMM4H} dataset, we compare with the following baseline methods: (1) \textit{CNN}~\cite{kim2014convolutional}, CNN for text classification;
\textit{LSTM}~\cite{hochreiter1997long}, long short-term memory network; 
(3) \textit{CNN+Att}~\cite{huynh2016adverse}, using attentive pooling after CNN model;
(4) \textit{LSTM+Att}~\cite{zhou2016attention}, applying attention pooling to \textit{LSTM};
(5) \textit{MSA}~\cite{wu2019msa}, a multi-head self-attention based approach for ADR detection;
(6) \textit{BERT}~\cite{miftahutdinov2019kfu}, using BERT for ADR detection.
The results on the \textit{MIND} and \textit{SMM4H} datasets are respectively shown in Tables~\ref{table.performance3} and~\ref{table.performance4}.\footnote{The baselines are trained on a centralized data storage.}
From the results, we find the performance of our \textit{FedKD} approach consistently outperform 
all the baseline methods (e.g., 70.7\% v.s. 68.5\% AUC scores on \textit{MIND}).
The further t-test results also show the differences between \textit{FedKD} and other baseline methods are significant ($p<0.05$).
This is because our approach takes the advantage of the state-of-the-art pre-trained language models and allows the teacher and student models to collaboratively learn from each other, which are helpful for learning strong models.

\begin{table}[!t]
\resizebox{0.48\textwidth}{!}{
\begin{tabular}{lcccc}
 \Xhline{1.5pt}
\multicolumn{1}{c}{\textbf{Methods}} & \textbf{AUC} & \textbf{MRR} & \textbf{nDCG@5} & \textbf{nDCG@10} \\ \hline
EBNR                                 & 66.1$\pm$0.3  & 31.9$\pm$0.3  & 34.9$\pm$0.3  & 40.5$\pm$0.4   \\
DKN                                  & 65.2$\pm$0.3  & 31.5$\pm$0.3  & 34.1$\pm$0.2  & 39.8$\pm$0.3   \\
NPA                                  & 67.4$\pm$0.2  & 32.6$\pm$0.3  & 35.5$\pm$0.3  & 41.3$\pm$0.3   \\
NAML                                 & 67.4$\pm$0.2  & 32.5$\pm$0.2  & 35.4$\pm$0.2  & 41.2$\pm$0.2   \\
LSTUR                                & 67.9$\pm$0.3  & 33.0$\pm$0.3  & 35.9$\pm$0.3  & 41.8$\pm$0.3   \\
NRMS                                 & 68.2$\pm$0.2  & 33.4$\pm$0.2  & 36.3$\pm$0.1  & 42.1$\pm$0.2   \\
FIM                                  & 68.5$\pm$0.3  & 33.6$\pm$0.2  & 36.6$\pm$0.3  & 42.4$\pm$0.3   \\ \hline
FedKD$_4$                               & \textbf{71.0}$\pm$0.1  & \textbf{35.6}$\pm$0.1  & \textbf{38.9}$\pm$0.1  & \textbf{44.8}$\pm$0.1    \\
FedKD$_2$                               & 70.5$\pm$0.1  & 35.3$\pm$0.2  & 38.6$\pm$0.1  & 44.3$\pm$0.2  \\ \Xhline{1.5pt}
\end{tabular}
}

\caption{Performance of different methods on \textit{MIND}.} \label{table.performance3} 
\end{table}

\begin{table}[!t]
\centering
\resizebox{0.42\textwidth}{!}{
\begin{tabular}{lccc}
 \Xhline{1.5pt}
\textbf{Methods} & \textbf{Precision} & \textbf{Recall} & \textbf{Fscore} \\ \hline
CNN              & 48.3$\pm$1.0 & 52.1$\pm$1.1  & 50.2$\pm$0.6            \\
LSTM             & 49.6$\pm$1.2 & 50.5$\pm$0.8  & 50.0$\pm$0.7            \\
CNN+Att          & 48.3$\pm$0.9 & 53.0$\pm$0.9  & 50.5$\pm$0.5            \\
LSTM+Att         & 48.5$\pm$0.5 & 52.7$\pm$0.7  & 50.5$\pm$0.4            \\
MSA              & 51.2$\pm$0.6 & 53.8$\pm$0.8  & 52.4$\pm$0.4            \\
BERT             & 56.6$\pm$0.6 & 59.8$\pm$0.9  & 58.0$\pm$0.6            \\ \hline
FedKD$_4$           & \textbf{59.4}$\pm$0.6 & \textbf{62.8}$\pm$0.9  & \textbf{60.7}$\pm$0.5         \\
FedKD$_2$           & 58.2$\pm$0.7 & 62.4$\pm$0.9  & 59.8$\pm$0.6        \\  \Xhline{1.5pt}
\end{tabular}
}

\caption{Performance of different methods on \textit{SMM4H}.} \label{table.performance4} 
\end{table}

\subsection{Additional Results on SMM4H}

\begin{figure}[!t]
  \centering
      \includegraphics[width=0.36\textwidth]{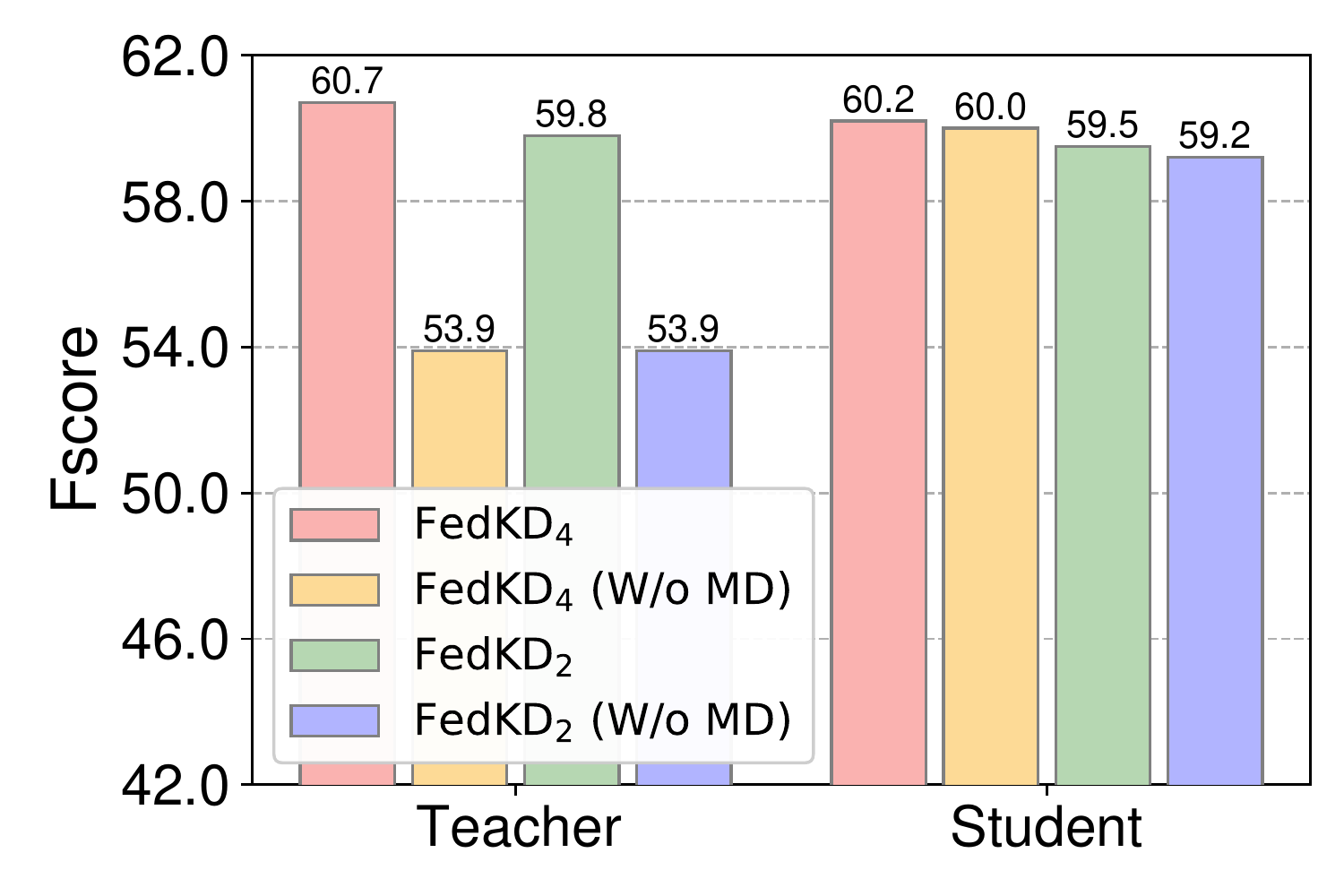}
  \caption{Influence of mutual distillation on the student and teacher models. }\label{fig.mutualq}

\end{figure}

\begin{figure}[!t]
  \centering
      \includegraphics[width=0.36\textwidth]{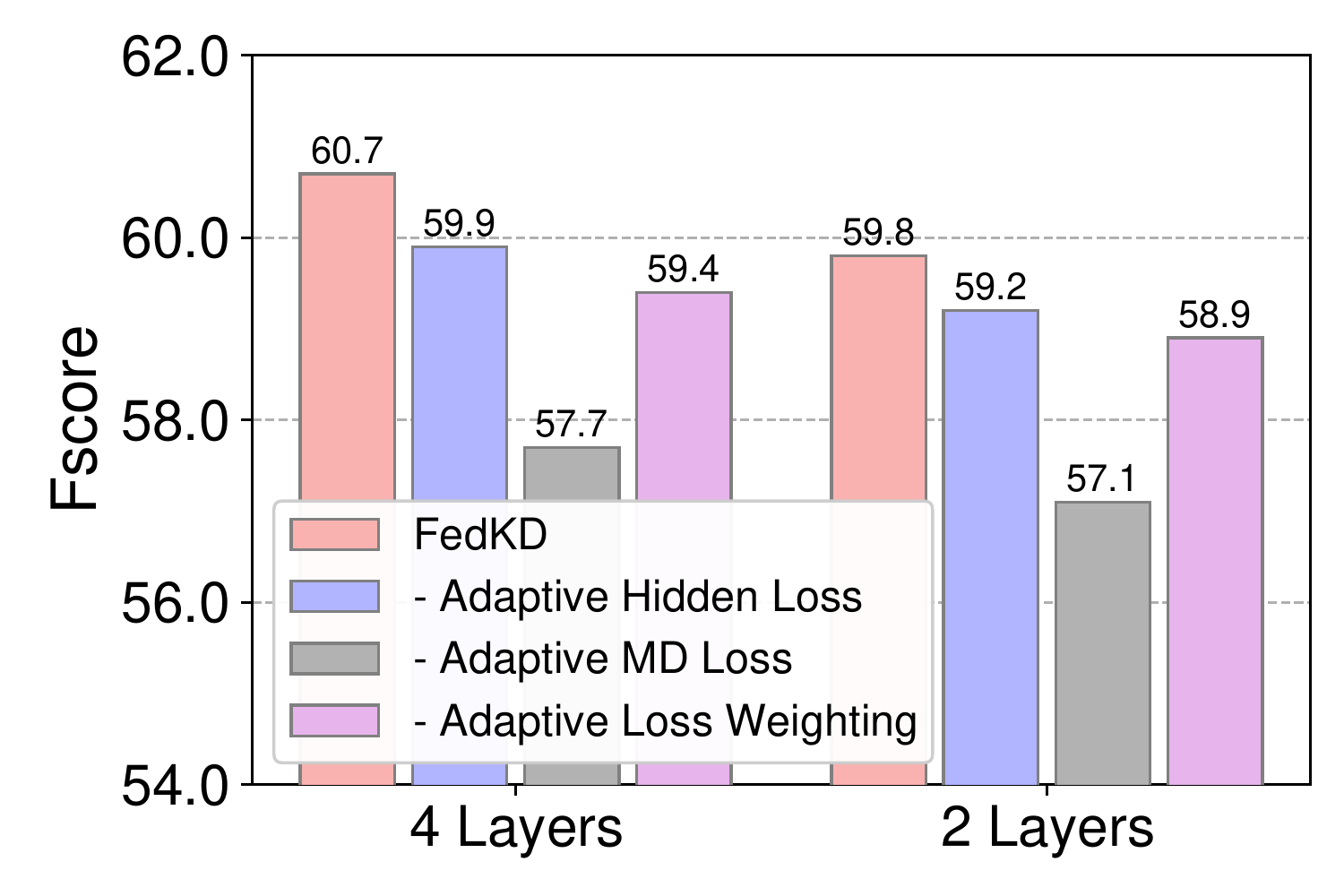}  
  \caption{Effect of adaptive mutual distillation. }\label{fig.ab2}

\end{figure}

\begin{figure}[!t]
  \centering
      \includegraphics[width=0.34\textwidth]{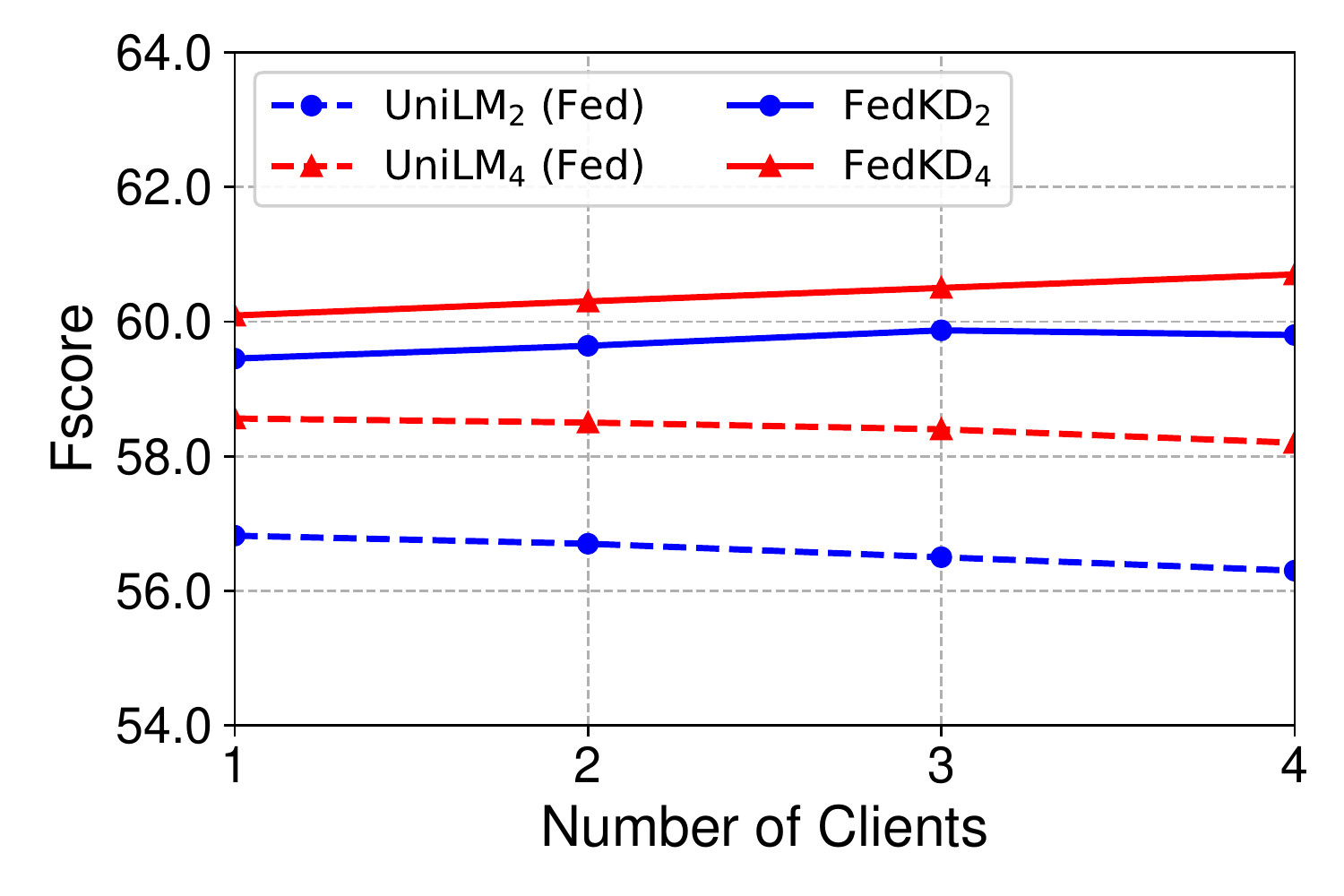}
  \caption{Influence of client number.}\label{fig.plat2}
\end{figure}

\begin{figure}[!t]
  \centering
  
  \subfigure[$T_{start}$.]{
    \includegraphics[width=0.223\textwidth]{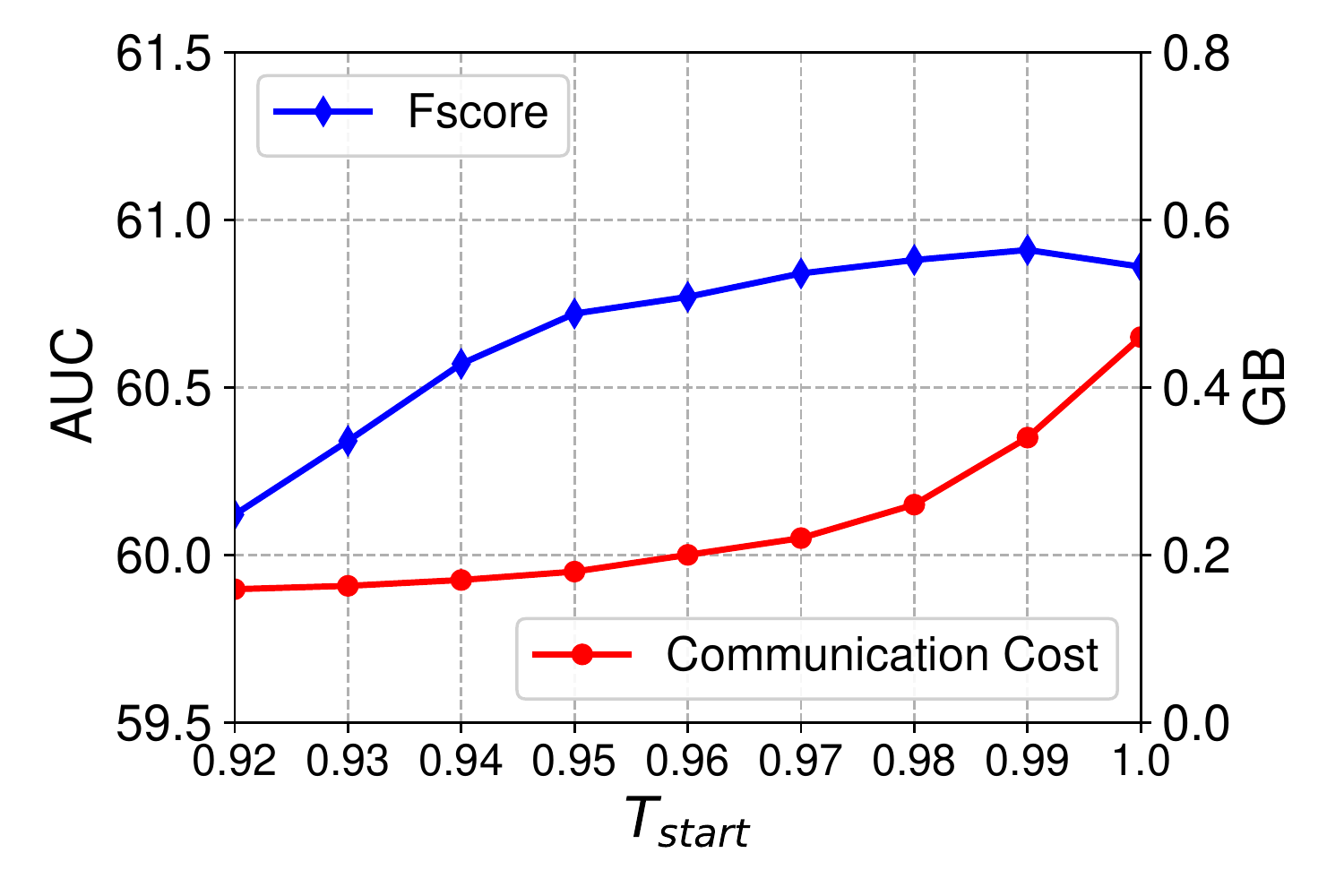}\label{fig.com2}
    }
      \subfigure[$T_{end}$.]{
    \includegraphics[width=0.223\textwidth]{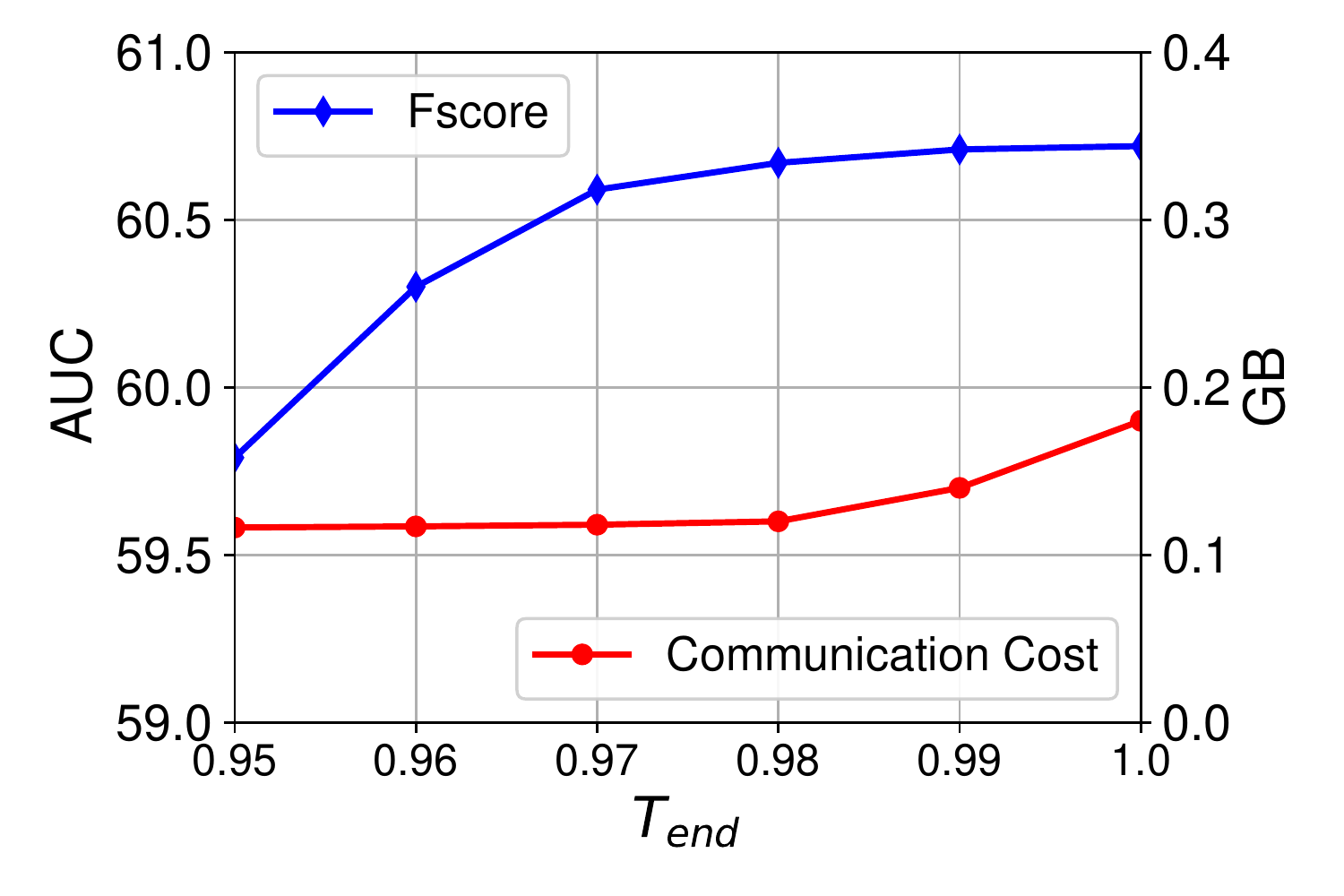}\label{fig.com4}
    }
  \caption{Influence of $T_{start}$ and $T_{end}$ on model performance and communication cost. }\label{fig.comt}

\end{figure}

We also report the additional results on the \textit{SMM4H} dataset, which are shown in Figs.~\ref{fig.mutualq}-\ref{fig.comt}.
We observe similar phenomena with the results on \textit{MIND}.

\subsection{Algorithm Workflow}

The workflow of \textit{FedKD} is summarized in Algorithm 1.
\begin{algorithm}[!t]
 \begin{algorithmic}[1]
     \STATE Setting the teacher learning rate $\eta_t$ and student learning rate $\eta_s$, client number $N$ 
     \STATE Setting the hyperparameters $T_{start}$ and $T_{end}$
      \FOR {each client $i$ (in parallel)} 
      \STATE Initialize parameters $\Theta^t_i$, $\Theta^s$
           
      \REPEAT
    \STATE $\mathbf{g}^t_i$,$\mathbf{g}_i$=\textbf{LocalGradients}($i$)
           \STATE $\Theta^t_i \leftarrow \Theta^t_i-\eta_t \mathbf{g}^t_i$
    \STATE  $\mathbf{g}_i \leftarrow \mathbf{U}_i\mathbf{\Sigma}_i\mathbf{V}_i$
    \STATE Clients encrypt $\mathbf{U}_i, \mathbf{\Sigma}_i, \mathbf{V}_i$
    \STATE Clients upload $\mathbf{U}_i, \mathbf{\Sigma}_i, \mathbf{V}_i$ to the server
    \STATE Server decrypts $\mathbf{U}_i, \mathbf{\Sigma}_i, \mathbf{V}_i$
    \STATE Server reconstructs $\mathbf{g}_i$
    
        \STATE Global gradients $\mathbf{g} \leftarrow 0$
    \FOR {each client $i$ (in parallel)} 
    \STATE $\mathbf{g}$=$\mathbf{g}$+$\mathbf{g}_i$
    \ENDFOR 
    \STATE $\mathbf{g} \leftarrow \mathbf{U}\mathbf{\Sigma}\mathbf{V}$
     \STATE Server encrypts $\mathbf{U}, \mathbf{\Sigma}, \mathbf{V}$
    \STATE Server distributes $\mathbf{U}, \mathbf{\Sigma}, \mathbf{V}$ to user clients 
    \STATE Clients decrypt $\mathbf{U}, \mathbf{\Sigma}, \mathbf{V}$
    \STATE Clients reconstructs $\mathbf{g}$
    \STATE $\Theta^s \leftarrow \Theta^s-\eta_s \mathbf{g}/N$
      \UNTIL{Local models converges}
       \ENDFOR
      \\[1.0ex]
    \textbf{LocalGradients}($i$):\\
          \STATE Compute task losses $\mathcal{L}^t_{t,i}$ and $\mathcal{L}^t_{s,i}$ 
    \STATE Compute losses $\mathcal{L}^d_{t,i}$, $\mathcal{L}^d_{s,i}$, $\mathcal{L}^h_{t,i}$, and $\mathcal{L}^h_{s,i}$
     \STATE  $\mathcal{L}^t_i \leftarrow \mathcal{L}^t_{t,i}+\mathcal{L}^d_{t,i}+\mathcal{L}^h_{t,i}$
          \STATE  $\mathcal{L}^s_i \leftarrow \mathcal{L}^t_{s,i}+\mathcal{L}^d_{s,i}+\mathcal{L}^h_{s,i}$
              \STATE Compute local teacher gradients $\mathbf{g}^t_i$ from $\mathcal{L}^t_i$
    \STATE Compute local student gradients $\mathbf{g}_i$ from $\mathcal{L}^s_i$
    
    \STATE \rm \textbf{return} $\mathbf{g}^t_i, \mathbf{g}_i$
     \\[1.0ex]
     
\end{algorithmic}
    \caption{FedKD}
\label{alg}
\end{algorithm}

\subsection{Experimental Environment}

Our experimental environment is built on a Linux server with Ubuntu 16.04 operation system.
The version of Python is 3.6
The server has 4 Tesla V100 GPUs with 32GB memory.
The CPU type is Intel(R) Xeon(R) Platinum 8168 CPU @ 2.70GHz.
The total memory is 128GB.
We use the horovod framework for parallel model training on the 4 GPUs, each of which represents a platform.

\subsection{Model Initialization}

In our approach, we use the token embedding layer and the first 4 or 2 layers of UniLM to initialize the student model.
We do not change the hidden dimension of the model because the  UniLMv2 models with other hidden dimensions are not released.
Note that our approach does not have limitations on the hidden dimension of the student model.

\subsection{Running Time}

On the \textit{MIND} dataset, the total training of time \textit{FedKD}$_{4}$ and \textit{FedKD}$_{2}$ are take around 66 and 57 hours, respectively.
On the \textit{SMM4H} dataset, their total training times are about 12 minutes and 10.5 minutes, respectively.

\subsection{Hyperparameter Settings}

The complete hyperparameter settings are listed in Table~\ref{hyper}.
The negative sampling ratio means the number of negative samples packed with each positive sample for model training.
We use the crossentropy loss to classify which sample is the positive sample.
The over sampling ratio means the repeating number of positive samples due to the high class imbalance. 
\begin{table}[h]
\centering
\resizebox{1.0\linewidth}{!}{
\begin{tabular}{l|c|c}
\hline
\multicolumn{1}{c|}{\textbf{Hyperparameters}}& \textit{MIND} & \textit{SMM4H}\\ \hline
LM hidden dimension                     & 768   & 768            \\ 
CNN feature map dimension                     & 256   & 256            \\ 
LSTM hidden dimension                     & 256   & 256            \\ negative sampling ratio                     & 4   & -            \\ 
over sampling ratio                     & -   & 2 for LMs, 9 for others            \\ 
attention query dimension                     & 200 & 200              \\ 
dropout                                      & 0.2 & 0.2             \\
optimizer                                    & Adam & Adam             \\
teacher model learning rate                                 & 2e-6         & 2e-6     \\
student model learning rate                                & 5e-6        & 1e-5      \\
batch size                                   & 32 & 64      \\   
$T_{start}$ & 0.95 & 0.95 \\  
$T_{end}$ & 0.98 & 0.98 \\  
Epoch & 3 & 2 \\
\hline
\end{tabular}
}
\caption{Hyperparameter settings.}\label{hyper}
\end{table}
 
\end{document}